\documentclass[conference]{IEEEtran}

\usepackage[T1]{fontenc}
\usepackage{amsmath,amssymb}
\usepackage{graphicx}
\usepackage{booktabs}
\usepackage{array}
\usepackage{multirow}
\usepackage{cite}
\usepackage[table]{xcolor}
\usepackage[hidelinks]{hyperref}
\usepackage{url}
\usepackage{enumitem}
\usepackage{siunitx}
\usepackage{tikz}
\usepackage{pgfplots}
\pgfplotsset{compat=1.18}
\usepgfplotslibrary{groupplots}
\usetikzlibrary{arrows.meta,positioning,calc,fit,backgrounds,shapes.misc}

\definecolor{okBlue}{HTML}{0072B2}
\definecolor{okOrange}{HTML}{E69F00}
\definecolor{okGreen}{HTML}{009E73}
\definecolor{okVerm}{HTML}{D55E00}
\definecolor{okSky}{HTML}{56B4E9}
\definecolor{okPurple}{HTML}{CC79A7}
\definecolor{okGrey}{HTML}{6E6E6E}

\tikzset{
  >={Stealth[length=4pt]},
  blk/.style={draw=#1, fill=#1!8, rounded corners=2pt, line width=0.6pt,
              align=center, font=\footnotesize, inner sep=3pt},
  blk/.default=okBlue,
  lane/.style={draw=okGrey!50, rounded corners=4pt, dashed, line width=0.5pt},
  lanelabel/.style={font=\scriptsize\bfseries\sffamily, text=okGrey},
  gate/.style={draw=okVerm, fill=white, rounded corners=1pt, font=\scriptsize\sffamily,
               inner sep=1.5pt, line width=0.5pt, text=okVerm},
  arr/.style={->, line width=0.6pt, draw=black!70},
  note/.style={font=\scriptsize\itshape, text=black!70, align=left},
}

\makeatletter
\def\abstract{\normalfont\@IEEEabskeysecsize\bfseries\textit{\abstractname.}\ \relax\@IEEEgobbleleadPARNLSP}
\def\IEEEkeywords{\normalfont\@IEEEabskeysecsize\bfseries\textit{\IEEEkeywordsname:}\ \relax\@IEEEgobbleleadPARNLSP}
\makeatother

\newcolumntype{L}[1]{>{\raggedright\arraybackslash}p{#1}}

\begin{document}

\title{From Zero to OCR: A Data-Centric Framework\\for Unsupported Low-Resource Scripts}

\author{%
\IEEEauthorblockN{Faizan Iqbal$^{\ast}$}
\IEEEauthorblockA{ORCID: \href{https://orcid.org/0009-0002-8998-9347}{0009-0002-8998-9347}\\
Hugging Face: \href{https://huggingface.co/faizaniqbal}{faizaniqbal}}
\and
\IEEEauthorblockN{Haq Nawaz Malik}
\IEEEauthorblockA{ORCID: \href{https://orcid.org/0009-0003-1994-7640}{0009-0003-1994-7640}\\
Hugging Face: \href{https://huggingface.co/Omarrran}{Omarrran}}
\and
\IEEEauthorblockN{Hafsah Pervaiz}
\IEEEauthorblockA{ORCID: \href{https://orcid.org/0009-0007-9226-1273}{0009-0007-9226-1273}\\
Hugging Face: \href{https://huggingface.co/HafsahPervaiz}{HafsahPervaiz}}
}

\maketitle

\begin{NoHyper}\renewcommand{\thefootnote}{}\footnotetext{$^{\ast}$Lead author.}\end{NoHyper}

\begin{abstract}
Optical character recognition (OCR) remains inaccessible to many low-resource scripts because annotated document images are unavailable and synthetic training data can fail to preserve script-specific visual structure. We present a data-centric framework that builds end-to-end OCR for such scripts from only an unannotated text corpus and digital fonts, without manually annotating a single training image. A synthetic data engine renders word, line and page images with correct complex-script shaping and automatically derived labels; a YOLOv8 word detector and a convolutional recurrent recognizer trained with connectionist temporal classification learn from these images; and a stage-gate evaluation protocol isolates recognition error before detection is introduced. We demonstrate the framework on Kashmiri, a Perso-Arabic Nastaliq script with approximately 6.8 million speakers that none of six widely used OCR engines supports. Three line-level recognizers, including one transferred from Arabic, stalled between 41\% and 79\% character error rate (CER), motivating a word-level formulation; subsequent improvements came primarily from rendering fidelity, diacritic-preserving crop margins and tuned non-maximum suppression. The resulting system achieves 0.39\% CER on held-out words, 10.0\% on synthetic pages and 13.5\% on browser-rendered digital pages, and establishes reference results of 18.45\% on typeset book pages and 49.8\% on degraded historical documents. We publicly release the trained models, the inference code and a synthetic dataset of 285{,}864 Kashmiri text-line images, and we describe a step-by-step procedure for building OCR for other unsupported scripts.
\end{abstract}


\section{Introduction}
\label{sec:intro}

\IEEEPARstart{O}{ptical} character recognition (OCR) is the first step in making printed text searchable, editable and usable as training data for language technology. Systems ranging from Tesseract to recent document vision-language models~\cite{smith2007tesseract,wei2024got,bai2025qwen25vl} have made this possible for many languages. Kashmiri is not among them. The language has approximately 6.8 million speakers in India~\cite{census2011language}, and its textbooks and literature are printed in a Perso-Arabic Nastaliq script. As Table~\ref{tab:engine_gap} shows, six widely used OCR engines support related scripts such as Urdu, Persian and Uyghur, but none supports Kashmiri. A printed Kashmiri page, such as the one in Fig.~\ref{fig:textbook}, therefore cannot be searched or reused without manual transcription. This gap reflects a broader pattern in which language technology coverage follows existing resources~\cite{joshi2020state,bird2020decolonising}.

\begin{table}[t]
\centering
\caption{Kashmiri recognition support in widely used OCR engines, checked against vendor documentation in October 2026.}
\label{tab:engine_gap}
\footnotesize
\begin{tabular}{@{}llc@{}}
\toprule
\textbf{Engine} & \textbf{Related scripts listed} & \textbf{Kashmiri} \\
\midrule
Tesseract~\cite{smith2007tesseract}           & Urdu, Persian, Sindhi, Uyghur & No \\
PaddleOCR~\cite{du2020ppocr}                  & Arabic, Persian, Urdu, Uyghur & No \\
EasyOCR~\cite{jaidedai2020easyocr}            & Arabic, Persian, Urdu, Uyghur & No \\
Google Cloud Vision~\cite{gcv2026languages}   & Urdu (experimental)           & No \\
Amazon Textract~\cite{aws2026textract}        & Six Latin-script languages    & No \\
Azure AI Vision~\cite{azure2026languages}     & Urdu (printed text)           & No \\
\bottomrule
\end{tabular}
\end{table}

\begin{figure*}[!t]
\centering
\begin{tikzpicture}[node distance=4mm and 4mm]
  \node[blk=okGrey, minimum width=2.3cm, minimum height=0.9cm] (corpus) {Unannotated\\Kashmiri corpus};
  \node[blk=okGrey, minimum width=2.3cm, minimum height=0.9cm, below=2mm of corpus] (fonts) {4 Naskh and\\Nastaliq fonts};
  \node[blk=okBlue, minimum width=3.3cm, minimum height=2.0cm, right=8mm of $(corpus.east)!0.5!(fonts.east)$, anchor=west] (engine)
       {\textbf{Synthetic data engine}\\[1pt]\scriptsize HarfBuzz + Raqm shaping\\\scriptsize right-to-left layout\\\scriptsize boxes from glyph metrics};
  \node[blk=okSky, minimum width=3.0cm, right=10mm of engine.north east, anchor=north west, yshift=2mm] (words) {Word crops + Unicode};
  \node[blk=okSky, minimum width=3.0cm, below=1.5mm of words] (lines) {Line images + word boxes};
  \node[blk=okSky, minimum width=3.0cm, below=1.5mm of lines] (pages) {Page images + boxes};
  \node[blk=okGreen, minimum width=2.9cm, minimum height=0.8cm, right=12mm of words] (rec) {\textbf{KashmiriCRNN}\\\scriptsize trained with CTC};
  \node[blk=okGreen, minimum width=2.9cm, minimum height=0.8cm, right=12mm of pages] (det) {\textbf{YOLOv8 detector}\\\scriptsize trained on renderer boxes};
  \draw[arr] (corpus.east) -- ++(4mm,0) |- (engine.west);
  \draw[arr] (fonts.east) -- ++(4mm,0) |- (engine.west);
  \draw[arr] (engine.east) -- ++(5mm,0) |- (words.west);
  \draw[arr] (engine.east) -- ++(5mm,0) |- (lines.west);
  \draw[arr] (engine.east) -- ++(5mm,0) |- (pages.west);
  \draw[arr] (words.east) -- (rec.west);
  \draw[arr] (lines.east) -- ++(6mm,0) |- (det.west);
  \draw[arr] (pages.east) -- (det.west);
  \node[gate, above=0.5mm of engine.north] (gA) {Gate A: ink density $\geq 1\%$};
  \node[gate, above=0.5mm of rec.north] (gB) {Gate B: oracle CER $\leq 15\%$};
  \coordinate (inf) at ($(fonts.south west)+(0,-17mm)$);
  \node[blk=okGrey, minimum width=1.9cm, minimum height=1.05cm, anchor=north west] (page) at (inf) {Page image};
  \node[blk=okOrange, minimum width=2.08cm, minimum height=1.05cm, right=4.2mm of page] (lseg) {Line\\segmentation\\\scriptsize Otsu + projection};
  \node[blk=okOrange, minimum width=2.08cm, minimum height=1.05cm, right=4.2mm of lseg] (wdet) {Word detection\\\scriptsize YOLOv8, NMS\\\scriptsize IoU 0.45, conf 0.35};
  \node[blk=okOrange, minimum width=2.08cm, minimum height=1.05cm, right=4.2mm of wdet] (pad) {Padded crops\\\scriptsize $+30$\,px H\\\scriptsize $+20$\,px V};
  \node[blk=okOrange, minimum width=2.08cm, minimum height=1.05cm, right=4.2mm of pad] (recog) {Recognition\\\scriptsize KashmiriCRNN\\\scriptsize CTC decoding};
  \node[blk=okOrange, minimum width=2.08cm, minimum height=1.05cm, right=4.2mm of recog] (rtl) {RTL reassembly\\\scriptsize sort by $x$-centroid,\\\scriptsize descending};
  \node[blk=okGrey, minimum width=1.5cm, minimum height=1.05cm, right=4.2mm of rtl] (out) {Unicode\\text};
  \foreach \a/\b in {page/lseg,lseg/wdet,wdet/pad,pad/recog,recog/rtl,rtl/out} \draw[arr] (\a) -- (\b);
  \draw[arr, dashed, draw=okGreen] (det.south) -- ++(0,-3mm) -| (wdet.north);
  \draw[arr, dashed, draw=okGreen] (rec.east) -- ++(4mm,0) |- ($(recog.north)+(0,6mm)$) -- (recog.north);
  \node[gate, above=1.2mm of rtl.north] (gCD) {Gates C/D: end-to-end CER, $\Delta_{\mathrm{det}}$};
  \coordinate (bR) at (out.east |- gB.north);
  \coordinate (iL) at (page.west |- out.south);
  \begin{scope}[on background layer]
    \node[lane, fit=(corpus)(fonts)(engine)(words)(pages)(rec)(det)(gA)(gB)(bR), inner sep=2.5mm, label={[lanelabel, anchor=south west]north west:BUILD TIME: NO MANUAL ANNOTATION}] {};
    \node[lane, fit=(page)(lseg)(wdet)(pad)(recog)(rtl)(out)(gCD), inner sep=2.5mm, label={[lanelabel, anchor=south west]north west:INFERENCE}] {};
  \end{scope}
\end{tikzpicture}
\caption{Overview of the proposed framework. \emph{Top:} the synthetic data engine renders a text corpus into word, line and page images with renderer-derived labels, which are used to train the recognizer and the detector. \emph{Bottom:} at inference, a page is segmented into lines; words are detected, cropped with diacritic-preserving margins, recognized, and reassembled in right-to-left order. Orange tags denote the stage gates of Section~\ref{sec:gates}; dashed green arrows denote trained weights.}
\label{fig:framework}
\end{figure*}
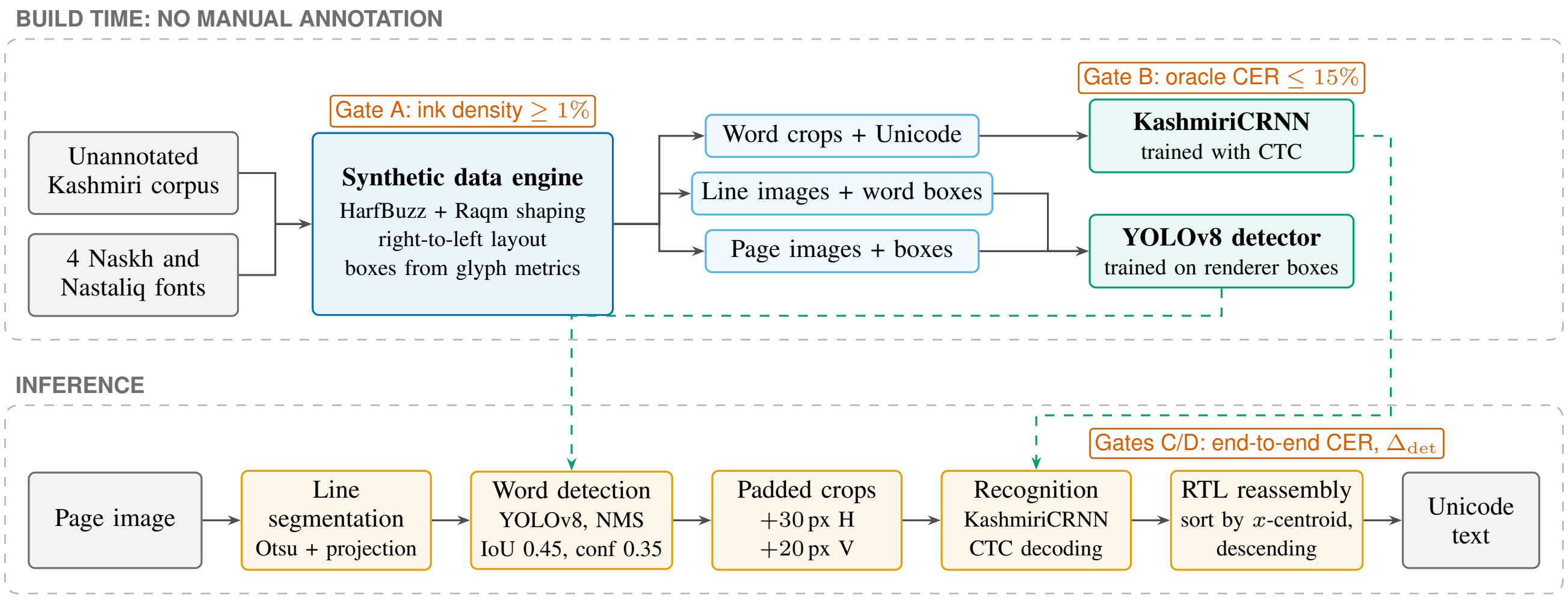

Kashmiri cannot be treated as a variant of Urdu. Its alphabet includes vowel letters and diacritics that Urdu does not use, and these marks distinguish otherwise identical words. This limits the most direct strategy, namely adapting a model trained on Urdu or Arabic, because such a model carries the language prior of its source. We hypothesize that an autoregressive decoder pre-trained on Urdu, when presented with an ambiguous glyph, tends to emit the Urdu sequence most frequent in its training data; we refer to this effect as \emph{linguistic cross-contamination}. Two observations are consistent with this concern. A recently published line-level Kashmiri baseline fine-tunes an Arabic Kraken model and reports that Kashmiri-specific diacritics remain difficult to recognize~\cite{fayaz2026kashmiri}. Our own initial model, a TrOCR model fine-tuned from an Arabic handwriting checkpoint, frequently misplaced the short-vowel marks \emph{zabar}, \emph{zer} and \emph{pesh}~\cite{iqbal2026koshurv3}. Neither observation isolates the cause, and Section~\ref{sec:limits} discusses this limitation. A second obstacle is the absence of training data: synthetic Kashmiri word images~\cite{malik2026ksocr} and small letter and word sets~\cite{jahan2025kashmiri} exist, but no annotated page images are available for learning word localization.

We therefore build the system from resources available for nearly every written language: running text and fonts capable of rendering it. Fig.~\ref{fig:framework} shows the resulting framework. The final design differs substantially from our initial approach. Our first recognizers operated on full text lines: a TrOCR model initialized from Arabic reached approximately 79\% CER, and a CRNN and PARSeq trained on synthetic Kashmiri lines converged at 41.2\% and 47.6\%. Several intermediate measurements also proved incorrect, including a 97.9\% CER caused by a rendering defect rather than by the recognizer. Because these findings shaped the method as much as the architecture did, we report them as part of the contribution.

The contributions of this paper are threefold. First, we describe a synthetic data engine that renders words, lines and pages with correct complex-script shaping and exports bounding boxes and Unicode labels automatically, together with a controlled experiment showing that the shaping path alone changes page CER from 35.2\% to 10.2\% (Section~\ref{sec:engine}). Second, we propose a stage-gate methodology that measures recognizer error on ground-truth boxes before a detector is introduced, and we present the evidence that motivated the move from line-level to word-level recognition (Section~\ref{sec:gates}). Third, we present an end-to-end Kashmiri OCR pipeline evaluated on six test sets, with ablations of crop padding and non-maximum suppression and an audit of six measurement errors (Sections~\ref{sec:system} to~\ref{sec:lessons}). The model weights and inference code are publicly released (Section~\ref{sec:release}).

\section{Related Work}
\label{sec:related}

\textbf{Text detection and recognition.} Most document and scene text systems separate text localization from transcription. Detectors such as EAST~\cite{zhou2017east}, CRAFT~\cite{baek2019craft} and YOLO~\cite{redmon2016yolo,jocher2023yolov8} propose text regions, which are then transcribed by CTC-trained convolutional recurrent networks~\cite{graves2006ctc,shi2017crnn}, attention-based decoders~\cite{shi2019aster} or transformers~\cite{li2023trocr,bautista2022parseq}. Page-level vision-language models such as Nougat~\cite{blecher2024nougat}, GOT-OCR2.0~\cite{wei2024got} and Qwen2.5-VL~\cite{bai2025qwen25vl} process entire documents but depend on large pre-training corpora in the scripts they support. We retain a decoupled design because, in the absence of real data, each stage must be trained on synthetic images and evaluated independently.

\begin{figure*}[!t]
\centering
\includegraphics[width=0.80\textwidth]{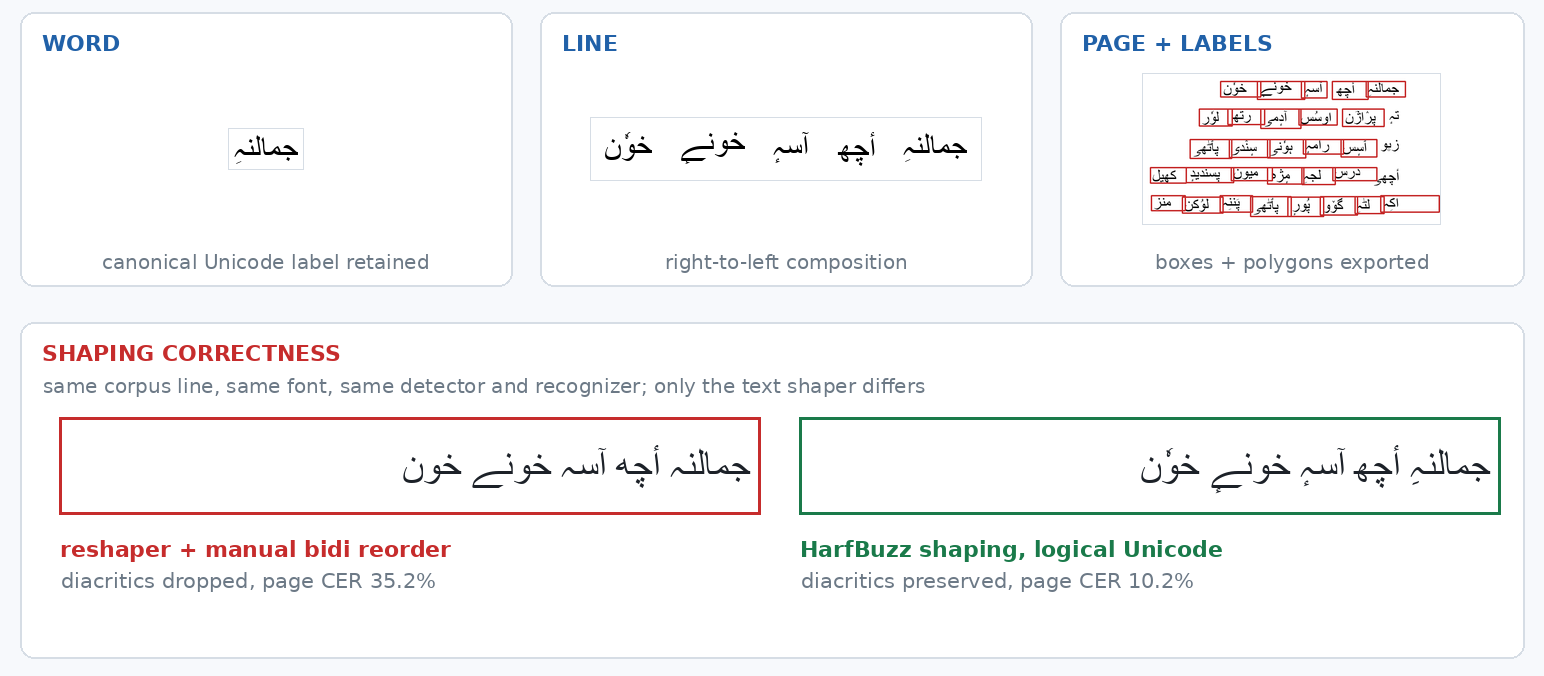}
\caption{Synthetic engine output and the effect of the shaping path. \emph{Top:} a single renderer call produces a word, line or page image with its canonical Unicode label and word boxes. \emph{Bottom:} the same corpus line rendered by two shaping paths, with font, detector, recognizer weights and thresholds held fixed. Presentation-form reshaping with manual bidirectional reordering (left) removes Kashmiri diacritics before rasterization and yields 35.2\% page CER; HarfBuzz shaping of logical Unicode (right) preserves them and yields 10.2\%.}
\label{fig:synthetic}
\end{figure*}

\textbf{Arabic-script and Nastaliq OCR.} Arabic-script text is written from right to left, letters take contextual forms, and adjacent characters combine into ligatures~\cite{lorigo2006arabic}. The Nastaliq style adds a baseline that descends diagonally within each word and diacritics placed above and below it, so that neighboring words overlap vertically~\cite{naz2014urdu}. For printed Urdu Nastaliq, line-level BLSTM recognizers trained on the UPTI corpus~\cite{sabbour2013upti,ulhasan2013urdu} and the high-resolution UTRNet model~\cite{rahman2023utrnet} demonstrate that segmentation-free recognition is effective when sufficient Urdu data are available. Such data do not exist for Kashmiri, and Section~\ref{sec:gates} reports the outcome of line-level recognition trained on synthetic Kashmiri text.

\textbf{Kashmiri OCR.} The 600K-KS-OCR dataset provides approximately 602{,}000 synthetic word images in three typefaces~\cite{malik2026ksocr}. Bashir \emph{et al.} discuss the challenges of recognizing Kashmiri script~\cite{bashir2022kashmiri}, and Jahan and Padmavathi compare CNN, SVM and $k$-NN classifiers on isolated Kashmiri letters and words~\cite{jahan2025kashmiri}. Fayaz \emph{et al.} fine-tune an Arabic Kraken model~\cite{kiessling2019kraken} on synthetic Kashmiri lines and report 79.91\% character accuracy~\cite{fayaz2026kashmiri}. Because their test data differ from ours, we do not compare results directly. To our knowledge, based on these works and the documentation summarized in Table~\ref{tab:engine_gap}, no openly released system converts a Kashmiri page image into Unicode text.

\textbf{Synthetic and data-centric training.} Synthetic rendering is an established method for training text recognizers~\cite{jaderberg2014synth,gupta2016synthtext}. The data-centric perspective treats the dataset as the primary object of engineering~\cite{zha2023datacentric}; DataComp, for example, fixes the training procedure and varies only data curation~\cite{gadre2023datacomp}. For an unsupported script, a more fundamental question precedes curation: whether the rendered glyphs are correct at all.

\section{Synthetic Data Engine}
\label{sec:engine}

The engine converts raw Kashmiri text into labeled images at word, line and page scale. It was also the source of the most consequential failures during development, none of which produced an error message.

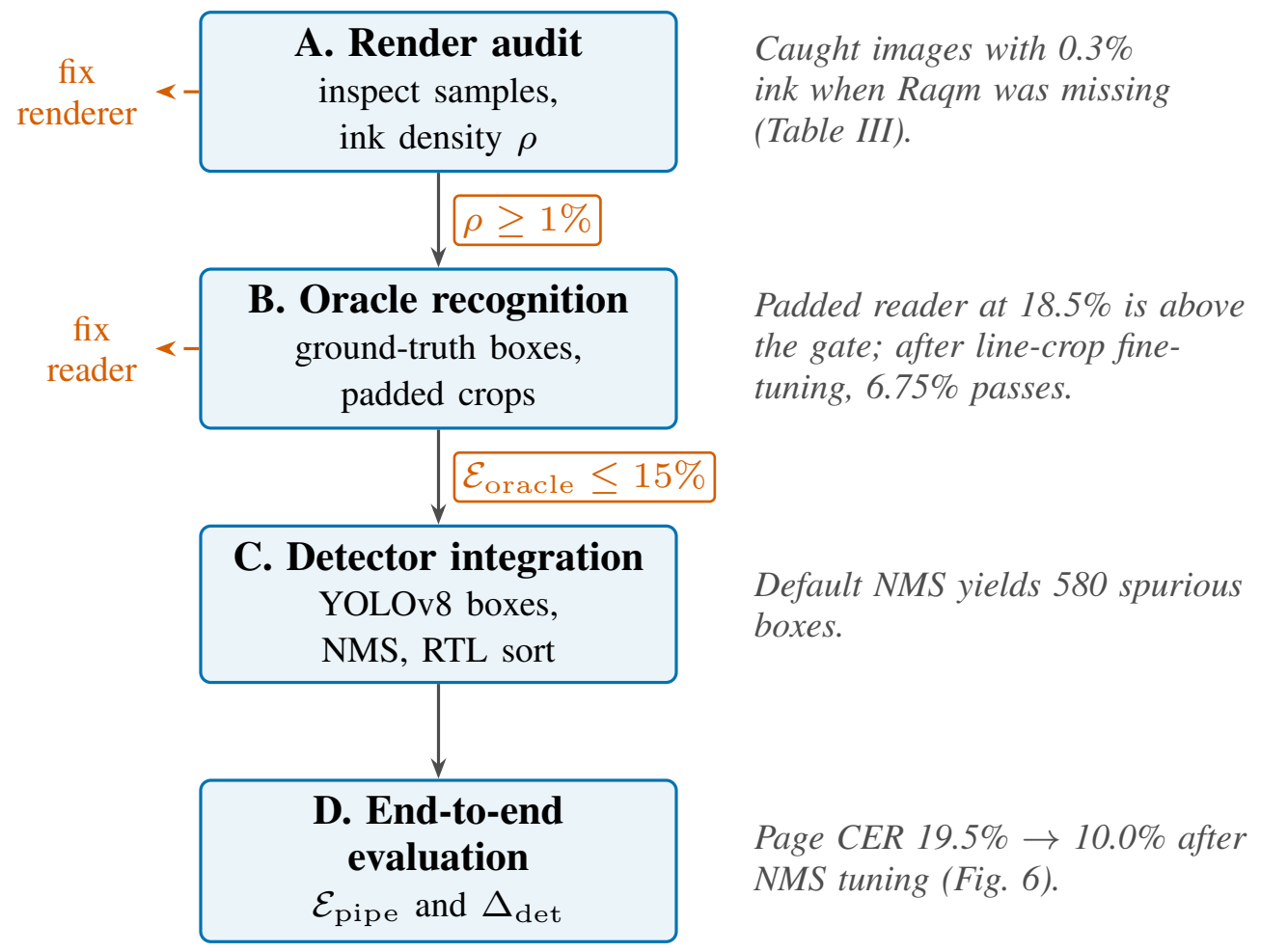
\begin{figure}[t]
\centering
\begin{tikzpicture}[node distance=6.5mm]
  \tikzset{stage/.style={blk=okBlue, text width=3.05cm, minimum height=0.9cm}}
  \node[stage] (A) {\textbf{A. Render audit}\\\scriptsize inspect samples, ink density $\rho$};
  \node[stage, below=of A] (B) {\textbf{B. Oracle recognition}\\\scriptsize ground-truth boxes, padded crops};
  \node[stage, below=of B] (C) {\textbf{C. Detector integration}\\\scriptsize YOLOv8 boxes, NMS, RTL sort};
  \node[stage, below=of C] (D) {\textbf{D. End-to-end}\\\textbf{evaluation}\\\scriptsize $\mathcal{E}_{\mathrm{pipe}}$ and $\Delta_{\mathrm{det}}$};
  \draw[arr] (A) -- node[gate, right=1mm] {$\rho\geq1\%$} (B);
  \draw[arr] (B) -- node[gate, right=1mm] {$\mathcal{E}_{\mathrm{oracle}}\leq15\%$} (C);
  \draw[arr] (C) -- (D);
  \node[note, text width=3.45cm, right=4mm of A] {Caught images with 0.3\% ink when Raqm was missing (Table~\ref{tab:audit}).};
  \node[note, text width=3.45cm, right=4mm of B] {Padded reader at 18.5\% is above the gate; after line-crop fine-tuning, 6.75\% passes.};
  \node[note, text width=3.45cm, right=4mm of C] {Default NMS yields 580 spurious boxes.};
  \node[note, text width=3.45cm, right=4mm of D] {Page CER 19.5\% $\rightarrow$ 10.0\% after NMS tuning (Fig.~\ref{fig:ablations}).};
  \draw[arr, okVerm, dashed] (A.west) -- ++(-3mm,0) node[left, font=\scriptsize, text=okVerm, align=center] {fix\\renderer};
  \draw[arr, okVerm, dashed] (B.west) -- ++(-3mm,0) node[left, font=\scriptsize, text=okVerm, align=center] {fix\\reader};
\end{tikzpicture}
\caption{Stage-gate methodology. Each stage must satisfy its gate (orange) before the next stage is built; a failed gate returns the pipeline to repair. Annotations on the right record the outcome observed at each gate during development.}
\label{fig:gates}
\end{figure}

\textbf{Corpus processing and shaping.} The corpus is cleaned by removing characters outside the Kashmiri inventory, normalizing whitespace and standardizing UTF-8 encoding. Correct rendering then requires contextual shaping. Let $U=(c_1,\dots,c_n)$ denote a sequence of logical Unicode characters. A shaping function $\mathcal{S}$ maps $U$ to a sequence of positioned presentation glyphs
\begin{equation}
  V = \mathcal{S}(U,\mathbf{F}),
  \label{eq:shaping}
\end{equation}
where $\mathbf{F}$ denotes the font metrics and OpenType tables. Rendering is performed through Pillow with Raqm~\cite{raqm} and HarfBuzz~\cite{harfbuzz}, which select isolated, initial, medial and final forms, position diacritics and apply right-to-left ordering. Four fonts are used. In the released line dataset, the two Nastaliq fonts, Gulmarg Nastaleeq and Noto Nastaliq Urdu, each account for one third of the lines, and the two Naskh fonts, Afan Koshur Naksh and Narqalam, for one sixth each~\cite{iqbal2026synth}. Training images are augmented with rotation, Gaussian blur, contrast jitter, line noise and character-spacing variation.

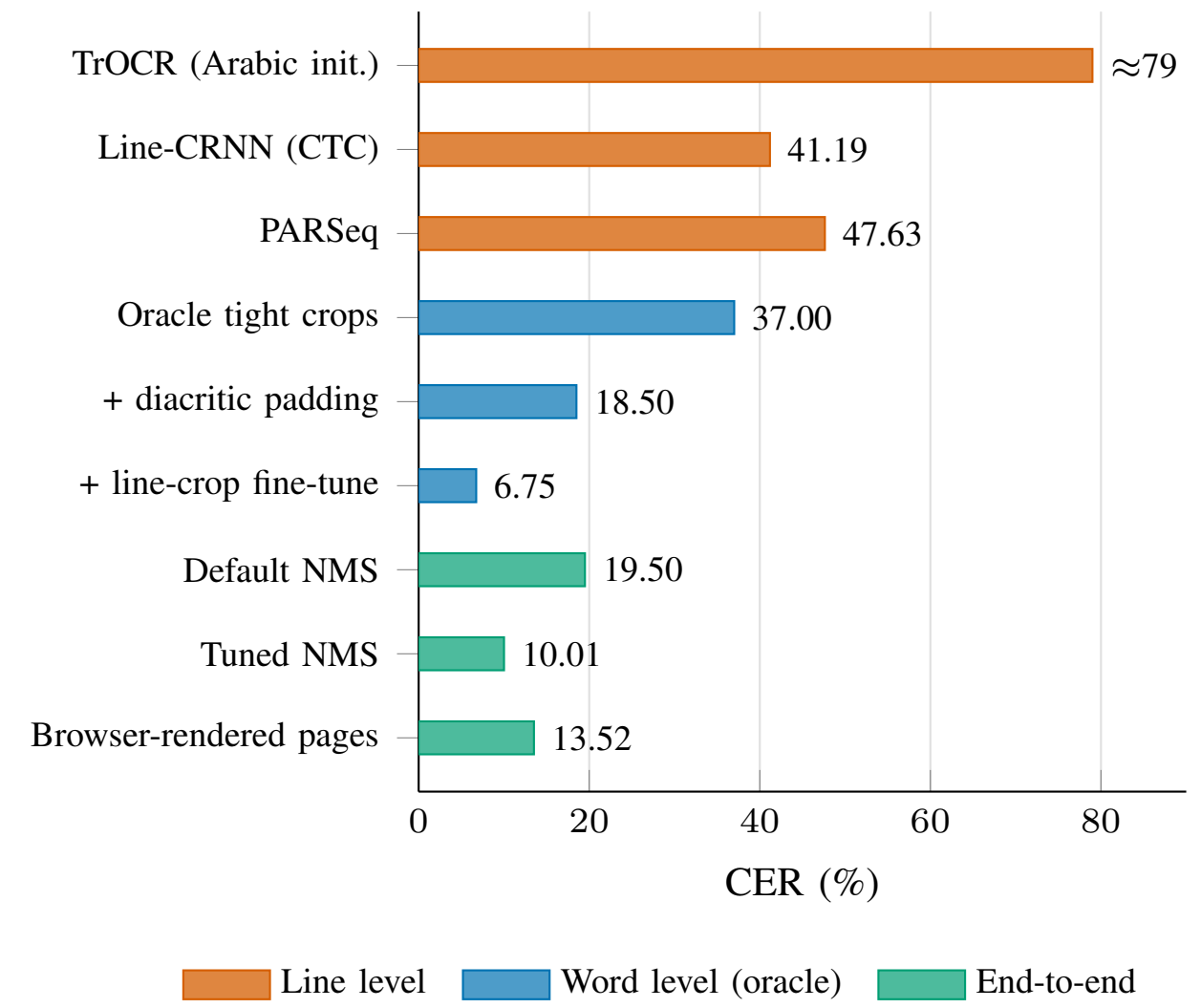
\begin{figure}[t]
\centering
\begin{tikzpicture}
\begin{axis}[
  xbar, scale only axis, width=0.6\columnwidth, height=5.4cm,
  bar width=6.5pt, bar shift=0pt, xmin=0, xmax=90,
  xlabel={CER (\%)}, xlabel style={font=\footnotesize},
  symbolic y coords={P9,P8,P7,P6,P5,P4,P3,P2,P1},
  ytick={P9,P8,P7,P6,P5,P4,P3,P2,P1}, y dir=reverse,
  yticklabels={TrOCR (Arabic init.), Line-CRNN (CTC), PARSeq, Oracle tight crops, + diacritic padding, + line-crop fine-tune, Default NMS, Tuned NMS, Browser-rendered pages},
  yticklabel style={font=\scriptsize}, xticklabel style={font=\scriptsize},
  nodes near coords, nodes near coords style={font=\scriptsize, anchor=west},
  point meta=explicit symbolic,
  axis lines*=left, xmajorgrids, grid style={black!12},
  enlarge y limits=0.08,
  legend style={font=\scriptsize, at={(0.32,-0.2)}, anchor=north, draw=none, legend columns=3, /tikz/every even column/.append style={column sep=5pt}}, area legend,
  legend cell align=left,
]
\addplot[fill=okVerm!75, draw=okVerm] coordinates {(79,P9)[$\approx$79] (41.19,P8)[41.19] (47.63,P7)[47.63]};
\addplot[fill=okBlue!70, draw=okBlue] coordinates {(37.00,P6)[37.00] (18.50,P5)[18.50] (6.75,P4)[6.75]};
\addplot[fill=okGreen!70, draw=okGreen] coordinates {(19.50,P3)[19.50] (10.01,P2)[10.01] (13.52,P1)[13.52]};
\legend{Line level, Word level (oracle), End-to-end}
\end{axis}
\end{tikzpicture}
\caption{Development trajectory. Colors denote measurement regimes, and CER values are comparable only within a regime. Line-level recognizers converge between 41\% and 79\% CER (the TrOCR value is taken from its model card); word-level recognition with padding and fine-tuning reaches 6.75\%; NMS tuning reduces synthetic-page CER to 10.0\%. The final bar corresponds to a different test set of browser-rendered digital pages.}
\label{fig:trajectory}
\end{figure}

\textbf{Automatic label generation.} During line and page composition, the engine records the bounding box of each word from the extents of its shaped glyphs. Each sample is stored with normalized boxes for detector training and the canonical logical Unicode string for recognizer training. Labels are stored as logical characters rather than presentation forms, since a recognizer trained on presentation forms produces output that displays correctly but does not match logical reference strings. The recognizer is trained on approximately 673{,}000 word and line crops derived from 600K-KS-OCR~\cite{malik2026ksocr}, and 285{,}864 rendered line images are released as Koshur-OCR-Synth~\cite{iqbal2026synth}. Fig.~\ref{fig:synthetic} (top) shows the engine output at all three scales.

\textbf{Rendering failure modes.} Two silent failures affected early versions of the engine. In the first, the Raqm library was absent from the build environment, and Pillow reverted to basic layout without warning; the resulting images contained only 0.3\% ink, and the recognizer reached 92.5\% CER on them. In the second, text was prepared with a standard Arabic reshaper followed by manual bidirectional reordering, a common approach for rendering Arabic in Python. This procedure removes Kashmiri diacritics before rasterization, so the recognizer was evaluated against marks that were never drawn. The lower panel of Fig.~\ref{fig:synthetic} isolates this effect: with the corpus, font, detector, recognizer and thresholds held fixed, changing only the shaping path changes page CER from 35.2\% to 10.2\%. Neither failure interrupted the pipeline, and both were identified only through visual inspection of the rendered images. This observation motivates the first gate of the methodology described next.

\section{Stage-Gate Methodology}
\label{sec:gates}

The methodology requires each stage of the pipeline to pass an explicit check before the next stage is built (Fig.~\ref{fig:gates}). Each gate corresponds to a failure that we encountered during development.

\textbf{Gate definitions.} Gate~A renders and inspects samples and halts if the mean ink density $\rho$ falls below 1\%, which indicates a broken shaping path. Gate~B crops every ground-truth word box with a margin of $(+30,+20)$ pixels, applies the recognizer, and halts if the resulting oracle CER $\mathcal{E}_{\mathrm{oracle}}$ exceeds 15\%. Only after both gates are passed do Stages~C and~D replace ground-truth boxes with detector output, sort the boxes from right to left, and measure the pipeline CER $\mathcal{E}_{\mathrm{pipe}}$. On a given evaluation set, the error attributable to detection is
\begin{equation}
  \Delta_{\mathrm{det}} = \mathcal{E}_{\mathrm{pipe}} - \mathcal{E}_{\mathrm{oracle}}.
  \label{eq:delta}
\end{equation}
Without Gate~B, a weak recognizer and a weak detector are indistinguishable at the pipeline level, since both produce a high page CER without indicating which component should be improved.

\textbf{Line-level recognition.} Our initial approach recognized complete lines, following prior work on Urdu~\cite{ulhasan2013urdu}. We first applied transfer learning: TrOCR~\cite{li2023trocr} was initialized from an Arabic handwriting checkpoint, its tokenizer was extended to cover all 47 Kashmiri letters, and it was fine-tuned on Kashmiri words and lines. This model reached approximately 79\% CER, dropping characters and misplacing short-vowel marks~\cite{iqbal2026koshurv3}. Line recognizers trained on synthetic Kashmiri lines performed better but remained inadequate. A line-level CRNN with CTC converged at 41.2\% CER; its training and validation losses were nearly equal (1.3270 and 1.3334), indicating that the model was underfitting rather than overfitting. PARSeq~\cite{bautista2022parseq}, which combines a vision-transformer encoder with an autoregressive decoder, converged at 47.6\% (Fig.~\ref{fig:trajectory}, line level). Had the architecture been the bottleneck, replacing it should have reduced the error substantially; instead, the error increased by 6.4 points. We therefore attributed the plateau to the task formulation. In Nastaliq, the baseline descends diagonally within each word and adjacent words overlap vertically, so a single left-to-right feature sequence over a full line represents the text poorly. We consequently adopted explicit word segmentation.

\textbf{Word-level recognition and crop margins.} Word-level recognition resolved the formulation problem but exposed a new one. With ground-truth boxes, tight crops produced 37.0\% CER. Inspection of the failed crops showed that Nastaliq diacritics, which lie above and below the baseline, fell outside the tight boxes. Expanding each box by 30 pixels horizontally and 20 pixels vertically halved the error to 18.5\%, which remained above the Gate~B threshold. Fine-tuning the recognizer on crops extracted from rendered lines reduced the error to 6.75\%, which satisfied the gate. With recognition no longer the dominant source of error, subsequent work focused on detection (Section~\ref{sec:ablations}).

\section{System Architecture}
\label{sec:system}

The deployed pipeline consists of four stages (Fig.~\ref{fig:framework}, bottom). Each stage is intentionally simple so that it can be verified in isolation.

\textbf{Line segmentation.} Pages are binarized using Otsu's method~\cite{otsu1979threshold}. The horizontal projection profile $H(y)=\sum_{x=1}^{W} I(x,y)$ is smoothed with a one-dimensional Gaussian kernel, and lines are separated at its minima. This procedure extracted 93\% of lines correctly in our development evaluation.

\textbf{Word detection.} A YOLOv8 detector~\cite{jocher2023yolov8} is trained on rendered lines with renderer-generated word boxes and achieves 99.6\% word recall in our development evaluation. Non-maximum suppression (NMS) removes overlapping candidates based on intersection over union, $\mathrm{IoU}(B_i,B_j)=|B_i\cap B_j|/|B_i\cup B_j|$. Because Nastaliq words overlap, the default thresholds (IoU 0.70, confidence 0.25) retain many duplicate boxes; we therefore use IoU 0.45 and confidence 0.35 (Section~\ref{sec:ablations}).

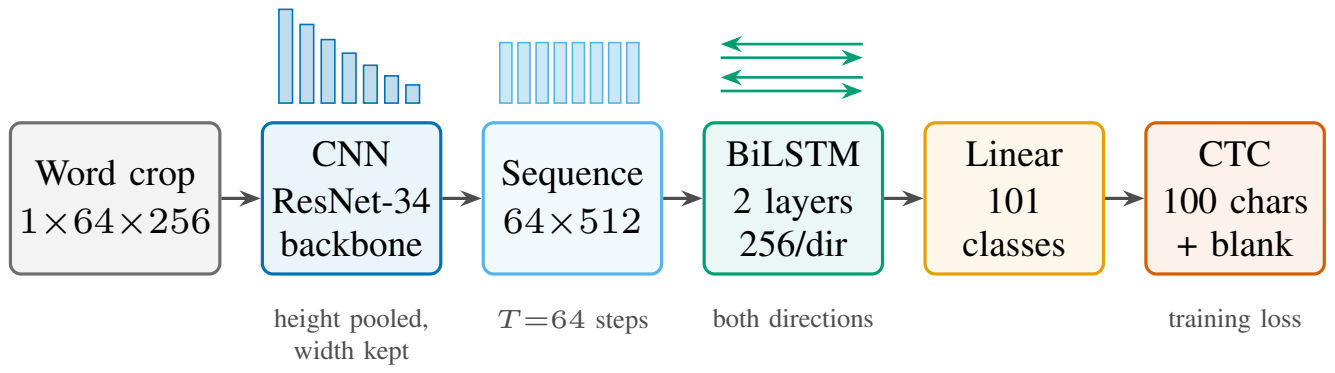
\begin{figure}[t]
\centering
\begin{tikzpicture}[node distance=2.6mm,
  mod/.style={blk=#1, minimum width=1.18cm, minimum height=1.0cm, font=\scriptsize, inner sep=1.5pt},
  dim/.style={font=\tiny, text=black!70, align=center}]
  \node[mod=okGrey]  (in)  {Word crop\\[1pt]$1{\times}64{\times}256$};
  \node[mod=okBlue, right=of in]  (cnn) {CNN\\[1pt]ResNet-34\\backbone};
  \node[mod=okSky, right=of cnn]  (seq) {Sequence\\[1pt]$64{\times}512$};
  \node[mod=okGreen, right=of seq] (rnn) {BiLSTM\\[1pt]2 layers\\256/dir};
  \node[mod=okOrange, right=of rnn] (lin) {Linear\\[1pt]101\\classes};
  \node[mod=okVerm, right=of lin] (ctc) {CTC\\[1pt]100 chars\\+ blank};
  \foreach \a/\b in {in/cnn,cnn/seq,seq/rnn,rnn/lin,lin/ctc} \draw[arr] (\a) -- (\b);
  \foreach \i/\h in {0/0.62,1/0.52,2/0.42,3/0.33,4/0.25,5/0.18,6/0.12}
    \draw[draw=okBlue, fill=okBlue!25, line width=0.3pt] ($(cnn.north west)+(0.12+\i*0.14,0.12)$) rectangle ++(0.09,\h);
  \foreach \k in {0,...,7}
    \draw[draw=okSky, fill=okSky!30, line width=0.3pt] ($(seq.north west)+(0.12+\k*0.12,0.12)$) rectangle ++(0.08,0.4);
  \foreach \j in {0,1} {
    \draw[->, okGreen, line width=0.5pt] ($(rnn.north west)+(0.12,0.2+\j*0.22)$) -- ++(0.94,0);
    \draw[<-, okGreen, line width=0.5pt] ($(rnn.north west)+(0.12,0.29+\j*0.22)$) -- ++(0.94,0);
  }
  \node[dim, below=1mm of cnn] {height pooled,\\width kept};
  \node[dim, below=1mm of seq] {$T{=}64$ steps};
  \node[dim, below=1mm of rnn] {both directions};
  \node[dim, below=1mm of ctc] {training loss};
\end{tikzpicture}
\caption{Architecture of the KashmiriCRNN recognizer (approximately 9.7M parameters). The convolutional backbone reduces the height of the crop while preserving its width, two bidirectional LSTM layers process the resulting 64-step feature sequence, and a linear layer assigns scores over 100 characters and the CTC blank at each step.}
\label{fig:crnn}
\end{figure}

\begin{table}[t]
\centering
\caption{CER and word exact-match accuracy by evaluation set. Exact match is reported for word-level sets only.}
\label{tab:results}
\footnotesize
\begin{tabular}{@{}llS[table-format=2.2]c@{}}
\toprule
\textbf{Level} & \textbf{Evaluation set} & {\textbf{CER (\%)}} & \textbf{Exact (\%)} \\
\midrule
\multirow{2}{*}{Word} & Synthetic word crops    & 2.11  & 94.2 \\
                      & Held-out word set        & 0.39  & 98.0 \\
\midrule
\multirow{4}{*}{Page} & Synthetic pages (4 fonts) & 10.00 & -- \\
                      & Browser-rendered digital  & 13.52 & -- \\
                      & Typeset book pages        & 18.45 & -- \\
                      & Degraded historical       & 49.80 & -- \\
\bottomrule
\end{tabular}
\end{table}

\textbf{Recognition.} Padded word crops are converted to grayscale, normalized to a height of 32 pixels, padded to $256\times64$, and transcribed by KashmiriCRNN (Fig.~\ref{fig:crnn}). A convolutional backbone derived from ResNet-34 maps each crop to a sequence of 64 feature vectors of dimension 512. A two-layer bidirectional LSTM with 256 hidden units per direction models context, and a linear layer projects each step onto 101 classes, comprising the 100 characters observed in the corpus and the CTC blank (Appendix~\ref{app:vocab}). Training minimizes the CTC loss~\cite{graves2006ctc}
\begin{equation}
  \mathcal{L}_{\mathrm{CTC}} = -\ln \sum_{\boldsymbol{\pi}\in\mathcal{B}^{-1}(\mathbf{y})} P(\boldsymbol{\pi}\mid\mathbf{x}),
  \label{eq:ctc}
\end{equation}
where $\mathcal{B}$ removes repeated labels and blanks. CTC was chosen over an autoregressive decoder deliberately: without a language decoder, the model has no Urdu or Arabic prior on which to rely. Training uses AdamW (learning rate $3\times10^{-4}$, weight decay $10^{-5}$) on NVIDIA Tesla T4 and RTX~3050 GPUs.

\textbf{Reassembly.} Word boxes within a line are sorted by horizontal centroid $x_{\min}+w/2$ in descending order, the recognized words are joined with spaces, and lines are output from top to bottom.

\section{Experiments}
\label{sec:experiments}

We first report accuracy on six evaluation sets, then examine the effect of individual design decisions, and finally analyze the remaining errors. Character error rate is defined as the total Levenshtein distance between predictions $\hat{y}_i$ and references $y_i$, divided by the total reference length:
\begin{equation}
  \mathrm{CER} = \frac{\sum_i \mathrm{Lev}(\hat{y}_i,y_i)}{\sum_i |y_i|}.
  \label{eq:cer}
\end{equation}
For word-level sets, exact-match accuracy is also reported.

\subsection{Main Results}
\label{sec:results}

Table~\ref{tab:results} summarizes the results. At the word level, the recognizer achieves 2.11\% CER on synthetic crops and 0.39\% on the held-out set, on which 98.0\% of words are transcribed exactly. End to end, the pipeline achieves 10.00\% CER on synthetic pages in four fonts and 13.52\% on browser-rendered digital pages (204 character edits over 1{,}509 reference characters across five pages). Performance degrades on physical print, with 18.45\% CER on pages from a typeset book and 49.80\% on degraded historical documents. We include these results because they indicate where synthetic training ceases to be sufficient; Section~\ref{sec:limits} discusses them further.

\subsection{Ablation Studies}
\label{sec:ablations}

\begin{figure}[t]
\centering
\begin{tikzpicture}
\begin{groupplot}[
  group style={group size=2 by 1, horizontal sep=1.1cm},
  width=0.545\columnwidth, height=4.3cm,
  ylabel style={font=\scriptsize}, xlabel style={font=\scriptsize},
  ticklabel style={font=\scriptsize}, title style={font=\scriptsize\bfseries, yshift=-1mm},
  axis lines*=left, ymajorgrids, grid style={black!12}, ymin=0,
]
\nextgroupplot[
  title={(a) Crop padding (oracle)},
  ybar, bar width=11pt, ymax=44,
  ylabel={CER (\%)},
  symbolic x coords={tight,H,HV}, xtick=data,
  xticklabels={{+0/+0},{+30/+0},{+30/+20}},
  xlabel={padding H/V (px)},
  nodes near coords={\pgfmathprintnumber[fixed,fixed zerofill,precision=1]{\pgfplotspointmeta}},
  nodes near coords style={font=\scriptsize},
  enlarge x limits=0.25,
]
\addplot[fill=okBlue!70, draw=okBlue] coordinates {(tight,37.0) (H,26.4) (HV,18.5)};
\nextgroupplot[
  title={(b) NMS thresholds (pages)},
  xmin=0, xmax=680, ymax=24,
  ylabel={end-to-end CER (\%)}, xlabel={spurious boxes},
  xtick={0,200,400,600},
]
\addplot[okGreen, line width=0.9pt, mark=*, mark size=2.2pt]
  coordinates {(580,19.5) (290,14.2) (120,10.0)};
\node[font=\tiny, anchor=north] at (axis cs:560,18.2) {0.70/0.25};
\node[font=\tiny, anchor=north west] at (axis cs:305,13.6) {0.55/0.30};
\node[font=\tiny, anchor=north west] at (axis cs:130,9.6) {0.45/0.35};
\end{groupplot}
\end{tikzpicture}
\caption{Ablation studies. (a)~Oracle word CER for three crop margins; adding the vertical margin reduces CER by a further 7.9 points. (b)~End-to-end CER against the number of spurious detections for three NMS settings, labeled as IoU/confidence thresholds.}
\label{fig:ablations}
\end{figure}
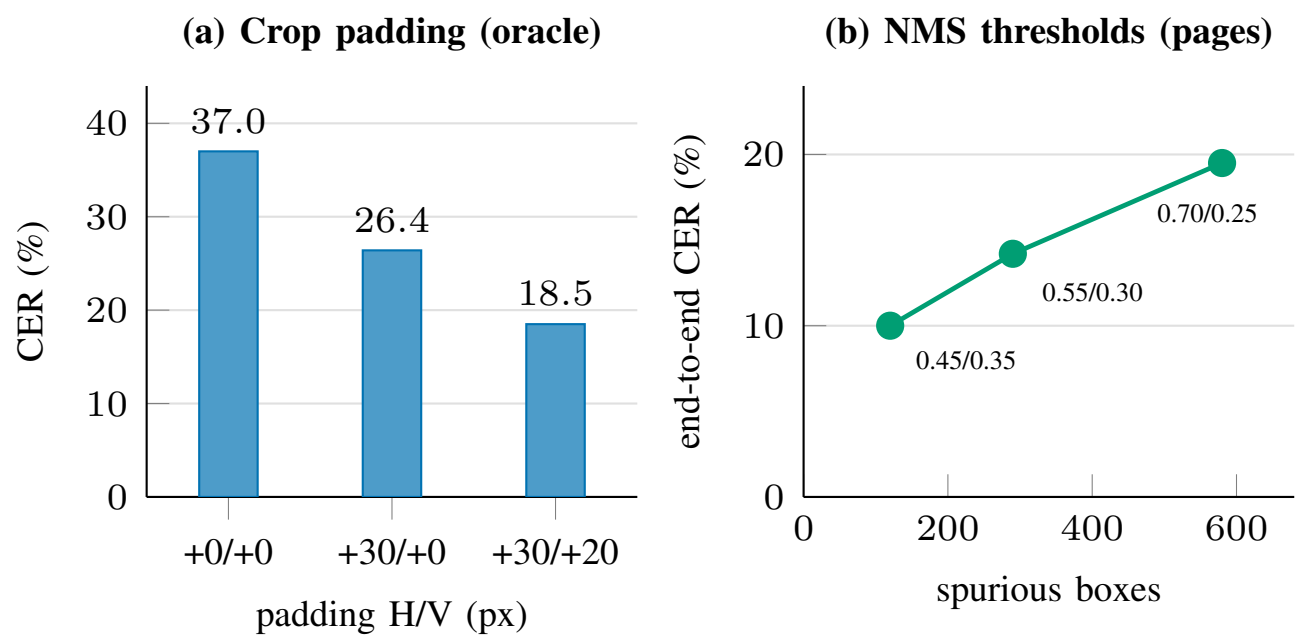

\textbf{Crop padding.} With ground-truth boxes and identical weights, horizontal padding reduces CER from 37.0\% to 26.4\%, and adding vertical padding reduces it further to 18.5\% (Fig.~\ref{fig:ablations}a). The vertical margin, which contains most Nastaliq diacritics, accounts for much of the remaining gain. Fig.~\ref{fig:errors}a shows a single box for which the margin changes an incorrect transcription into an exact one.

\textbf{NMS thresholds.} Lowering the IoU threshold from 0.70 to 0.45 and raising the confidence threshold from 0.25 to 0.35 reduces spurious boxes from 580 to 120 and end-to-end CER from 19.5\% to 10.0\% (Fig.~\ref{fig:ablations}b). Duplicate boxes over a single word produce repeated or fragmented transcriptions; once the recognizer satisfies Gate~B, this is the largest single improvement we observed, and it requires no retraining.

\textbf{Shaping.} Replacing HarfBuzz shaping of logical Unicode with presentation-form reshaping and manual reordering increases page CER from 10.2\% to 35.2\%, with all other components fixed (Fig.~\ref{fig:synthetic}). This effect exceeds that of either inference-time decision above, which indicates that the quality of a synthetic pipeline is bounded by the correctness of its rendering.

\begin{table}[t]
\centering
\caption{Measurement artifacts caught and corrected during development.}
\label{tab:audit}
\footnotesize
\begin{tabular}{@{}L{1.45cm}L{1.55cm}L{4.8cm}@{}}
\toprule
\textbf{Stage} & \textbf{Reported} & \textbf{Root cause} \\
\midrule
Rendering   & 97.9\% CER     & Arabic reshaper dropped Kashmiri diacritics \\
Rendering   & 92.5\% CER     & Raqm missing; images had 0.3\% ink \\
Evaluation  & 36.97\% CER    & Stale tight-crop value; padded CER was 18.5\% \\
End-to-end  & Inflated CER   & Wrong recognizer checkpoint loaded \\
Install     & Clean pass     & Inherited site packages hid a missing SciPy \\
GPU test    & Code failure   & Tesla P100 (sm\_60) incompatibility misread as a bug \\
\bottomrule
\end{tabular}
\end{table}

\subsection{Error Analysis and Qualitative Results}
\label{sec:qualitative}

\begin{figure*}[!t]
\centering
\includegraphics[width=0.98\textwidth]{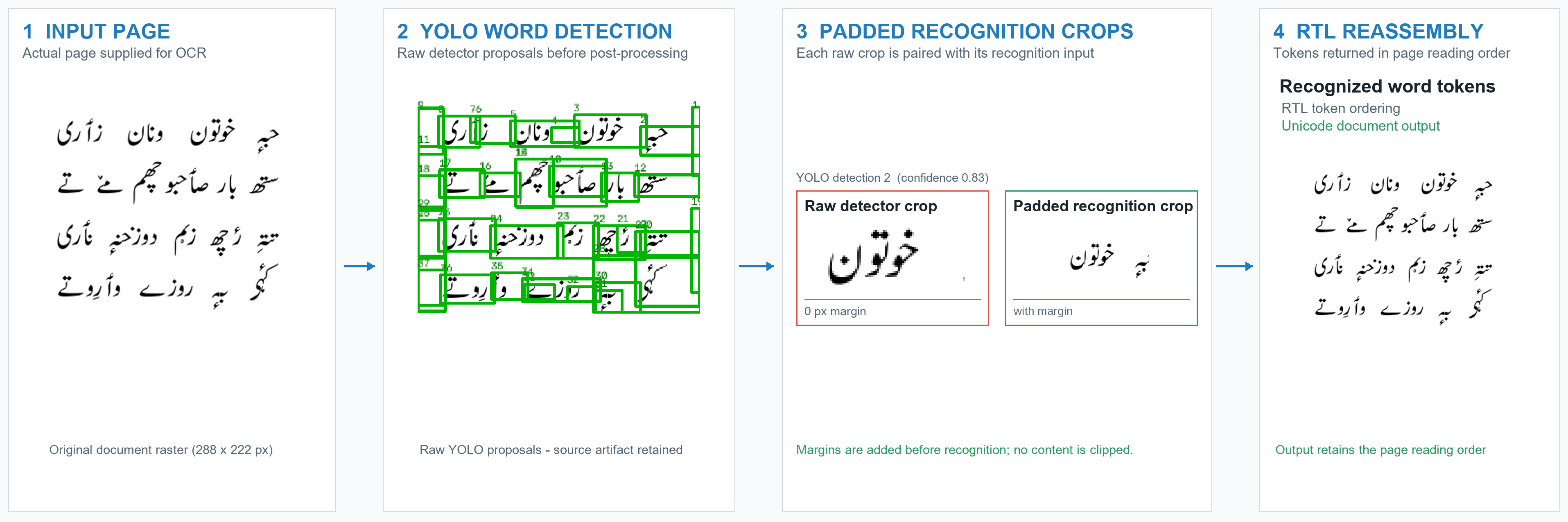}
\caption{End-to-end trace of a rendered Kashmiri page. (1)~Input page. (2)~Raw YOLOv8 proposals before NMS, in which overlapping words attract multiple boxes. (3)~A raw detector crop and the corresponding padded crop passed to the recognizer. (4)~Recognized words reassembled in right-to-left reading order.}
\label{fig:trace}
\end{figure*}

Fig.~\ref{fig:trace} traces a single page through the pipeline. Panel~2 shows the raw detector output before NMS and illustrates why NMS tuning is effective: overlapping Nastaliq words attract multiple boxes each.

\begin{figure*}[!t]
\centering
\includegraphics[width=0.76\textwidth]{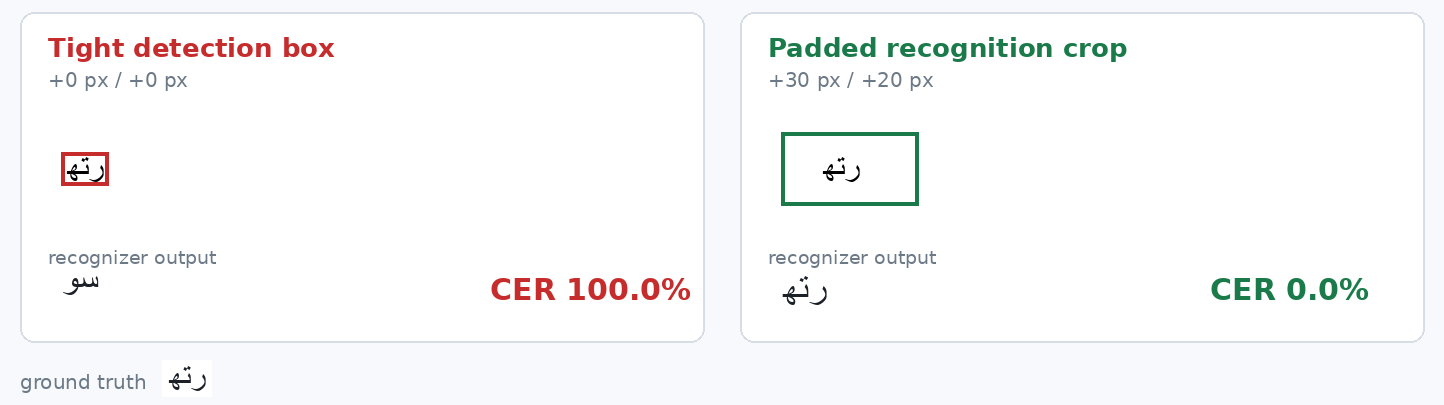}\\[1pt]
{\footnotesize (a) Crop margin}\\[5pt]
\includegraphics[width=0.76\textwidth]{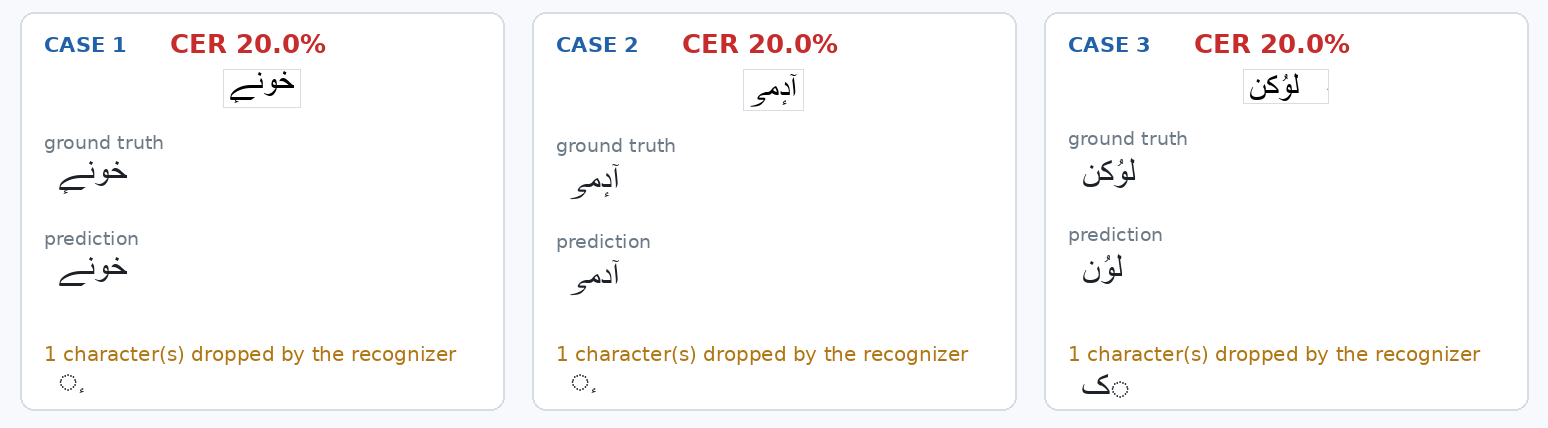}\\[1pt]
{\footnotesize (b) Residual errors}
\caption{Effect of crop margins and residual errors. (a)~A tight box clips the diacritics and the word is misrecognized; the padded crop of the same box is recognized exactly. (b)~The three highest-CER words from the run in Fig.~\ref{fig:trace}, selected automatically. Each contains one omitted character: two diacritics and one base letter.}
\label{fig:errors}
\end{figure*}

Fig.~\ref{fig:errors}b shows the three highest-CER words from the same run, selected automatically by error magnitude. Each contains a single deleted character, corresponding to 20\% CER on a five-character word; two are omitted diacritics and one is an omitted base letter. None involves the substitution of a Kashmiri character by an Urdu or Arabic one. Three cases are insufficient to establish a trend, but they are consistent with a recognizer that has no borrowed language prior. Across the 28 matched boxes in the run of Fig.~\ref{fig:errors}a, mean word CER decreases from 45.2\% with tight crops to 2.7\% with padded crops.

\begin{figure*}[!t]
\centering
\includegraphics[width=0.74\textwidth]{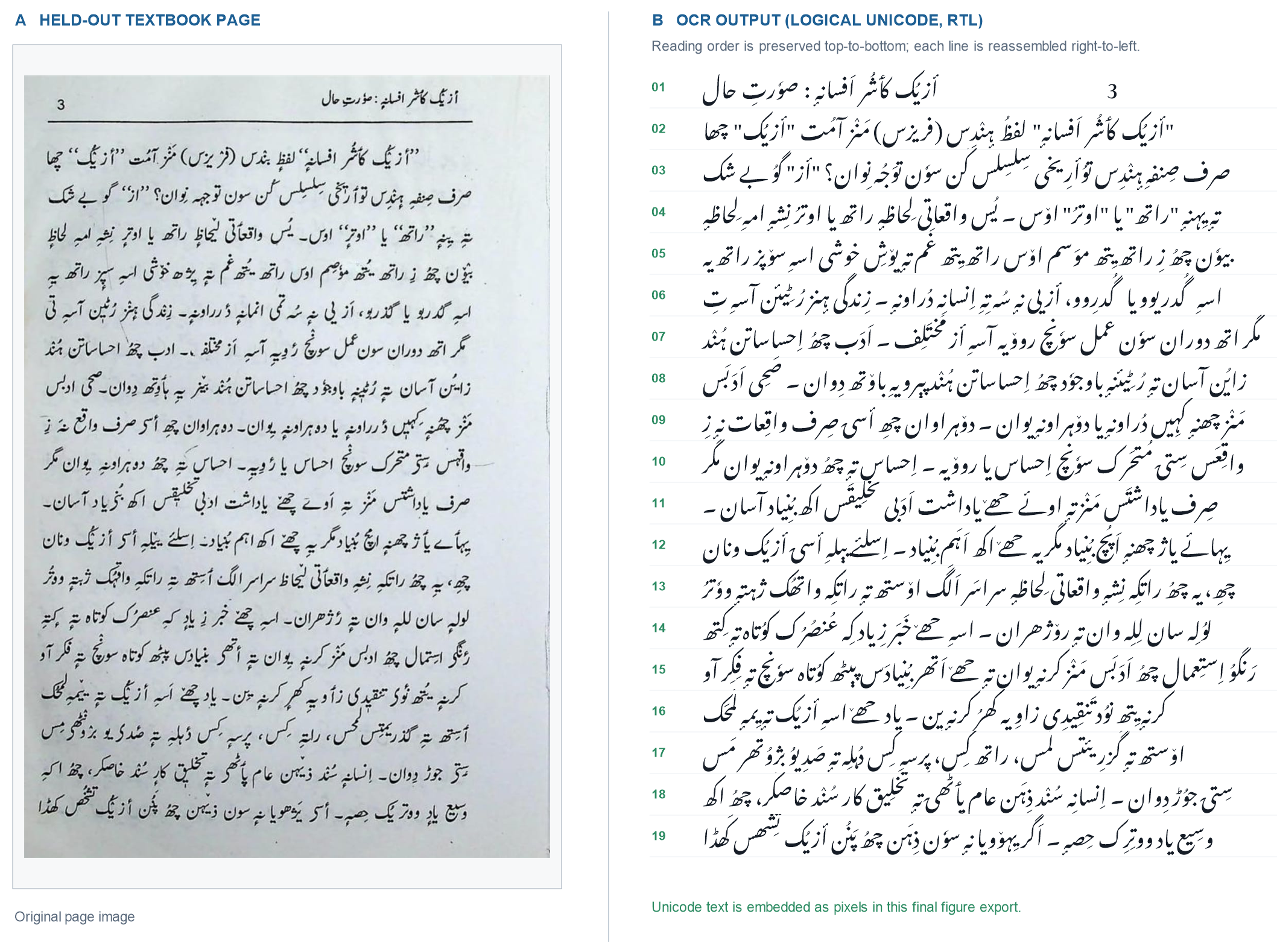}
\caption{Qualitative result on a held-out Kashmiri textbook page. \emph{Left:} the page image. \emph{Right:} the logical Unicode text produced by the pipeline, one line per detected text line, rendered for display with line numbers added. The example is illustrative; quantitative results on typeset book pages are reported in Table~\ref{tab:results}.}
\label{fig:textbook}
\end{figure*}

Fig.~\ref{fig:textbook} shows the output of the pipeline on a held-out page image from a Kashmiri textbook. Reading order, Kashmiri-specific letters and most diacritics are preserved. This example is intended to illustrate the output format rather than to quantify accuracy; the measured CER on typeset book pages is 18.45\% (Table~\ref{tab:results}).

\section{Discussion}
\label{sec:lessons}

\label{sec:audit}
Several findings from this work are specific to Kashmiri, but we expect most of them to apply to other unsupported scripts. We summarize them here.

\textbf{Visual inspection precedes metric evaluation.} Both rendering failures described in Section~\ref{sec:engine} completed without error and were detected only by inspecting the rendered images.

\textbf{Convergent failures suggest a task-level bottleneck.} The line-level plateau illustrates this principle. Once a second architecture converged to a similar error range, further tuning of either model was unlikely to be productive.

\textbf{Reported metrics require independent verification.} Six metrics reported during development were later found to be measurement artifacts rather than properties of the model (Table~\ref{tab:audit}). Two arose from rendering defects, one from a stale value carried forward, one from loading an incorrect checkpoint, and two from the software environment. Each was identified by inspecting the actual input image and recomputing CER from the raw reference and prediction strings. We now accept a metric only after both steps have been performed.

\begin{table}[t]
\centering
\caption{Artifacts and their release status.}
\label{tab:artifacts}
\footnotesize
\begin{tabular}{@{}L{2.55cm}L{3.9cm}l@{}}
\toprule
\textbf{Artifact} & \textbf{Content} & \textbf{Status} \\
\midrule
Recognizer weights & KashmiriCRNN, about 9.7M parameters & Released \\
Detector weights   & YOLOv8 word detector            & Released \\
Inference code     & \texttt{koshur\_ocr.py} CLI and Python API, runs locally; NMS post-processing & Released \\
Vocabulary         & Character map and tokenizer     & Released \\
Line dataset       & Koshur-OCR-Synth, 285{,}864 rendered lines, four fonts, Apache-2.0 & Released \\
Rendering engine   & Text and fonts to labeled word, line and page images & Planned \\
Benchmark          & Evaluation sets of Table~\ref{tab:results} & Planned \\
\bottomrule
\end{tabular}
\end{table}

\textbf{Geometric processing choices dominate residual error.} After the move to word-level recognition, the two largest improvements came from the crop margin and the NMS thresholds, neither of which involved changes to the network.

\section{Release and Replication}
\label{sec:release}

Table~\ref{tab:artifacts} lists the released and planned artifacts. The model repository is gated, and users must accept its access conditions before downloading.\footnote{\url{https://huggingface.co/Faizaniqbal/KoshurOCR-CRNN-Baseline}}

\textbf{Replication procedure.} The framework requires a text corpus and fonts for the target script. (1)~Collect unannotated text. (2)~Derive the character inventory from the corpus to define the output layer. (3)~Configure Pillow with Raqm and the script's fonts, and verify the shaping visually (Gate~A). (4)~Render lines and pages with exported boxes and Unicode labels. (5)~Train the recognizer and evaluate it on ground-truth crops (Gate~B). (6)~Add the detector, padded cropping and the script's reading order, and tune NMS on held-out pages (Gates~C and~D).

Several languages written in Perso-Arabic script, including Balochi, Brahui, Khowar and Shina, share the properties that shaped this design, namely contextual joining, dense diacritics and right-to-left order, and have no Tesseract model. Whether the thresholds reported here transfer to these scripts remains an open question.

\section{Limitations and Ethical Considerations}
\label{sec:limits}

\textbf{Domain gap.} All training data are synthetic. CER increases from 10.0\% on synthetic pages to 18.45\% on typeset book pages and 49.80\% on degraded historical documents (Table~\ref{tab:results}). The system is therefore not yet suitable for archival material without adaptation to real images.

\textbf{Evaluation scope.} The browser-rendered set contains five pages (1{,}509 characters), and the padding and NMS ablations each consist of a single sweep. We have not compared our system with the Kraken baseline~\cite{fayaz2026kashmiri} or with a fine-tuned vision-language model on a shared test set, and the 79\% TrOCR result is taken from its model card rather than measured on our test sets. The cross-contamination hypothesis stated in Section~\ref{sec:intro} therefore remains untested. The padding margins depend on rendering resolution and were tuned on our data.

\textbf{Layout.} Projection-based line segmentation assumes a single column of approximately horizontal lines and extracted 93\% of lines correctly. Multi-column pages, tables and marginalia are outside the current scope.

\textbf{Data, licensing and intended use.} The engine renders images only from its input corpus; the provenance and licenses of the corpus and fonts therefore determine how derived images may be shared, and both will be documented with the engine release. The model is released under the MIT license through a gated repository and is intended for digitizing and searching Kashmiri print. Given the error rates reported above, its output should not replace human transcription of cultural or legal documents.

\section{Conclusion}
\label{sec:conclusion}

We have presented a publicly available OCR system that converts Kashmiri print into Unicode text and is built from a text corpus and four fonts rather than from annotated images. The recognizer is a standard CRNN; the system's effectiveness derives from a sequence of measured findings. The text-shaping path determines whether synthetic training data are correct; convergent failures across architectures suggest that the task formulation, rather than the model, requires revision; and once recognition is reliable, crop margins and NMS thresholds account for most of the remaining improvement. Performance on printed and historical documents remains the principal limitation, and adaptation to real images is the focus of our ongoing work. We expect the stage-gate protocol, the replication procedure and the documented measurement errors to reduce the development effort required for other unsupported scripts.

\bibliographystyle{IEEEtran}
\bibliography{references}

\begin{thebibliography}{10}
\providecommand{\url}[1]{#1}
\csname url@samestyle\endcsname
\providecommand{\newblock}{\relax}
\providecommand{\bibinfo}[2]{#2}
\providecommand{\BIBentrySTDinterwordspacing}{\spaceskip=0pt\relax}
\providecommand{\BIBentryALTinterwordstretchfactor}{4}
\providecommand{\BIBentryALTinterwordspacing}{\spaceskip=\fontdimen2\font plus
\BIBentryALTinterwordstretchfactor\fontdimen3\font minus
  \fontdimen4\font\relax}
\providecommand{\BIBforeignlanguage}[2]{{%
\expandafter\ifx\csname l@#1\endcsname\relax
\typeout{** WARNING: IEEEtran.bst: No hyphenation pattern has been}%
\typeout{** loaded for the language `#1'. Using the pattern for}%
\typeout{** the default language instead.}%
\else
\language=\csname l@#1\endcsname
\fi
#2}}
\providecommand{\BIBdecl}{\relax}
\BIBdecl

\bibitem{smith2007tesseract}
R.~Smith, ``An overview of the {Tesseract} {OCR} engine,'' in \emph{Proc. 9th
  Int. Conf. Document Analysis and Recognition (ICDAR)}, 2007, pp. 629--633.

\bibitem{wei2024got}
H.~Wei, C.~Liu, J.~Chen, J.~Wang, L.~Kong, Y.~Xu, Z.~Ge, L.~Zhao, J.~Sun,
  Y.~Peng, C.~Han, and X.~Zhang, ``General {OCR} theory: Towards {OCR}-2.0 via
  a unified end-to-end model,'' arXiv:2409.01704, 2024.

\bibitem{bai2025qwen25vl}
S.~Bai \emph{et~al.}, ``{Qwen2.5-VL} technical report,'' arXiv:2502.13923,
  2025.

\bibitem{census2011language}
{Office of the Registrar General \& Census Commissioner, India}, ``{Census of
  India 2011, Paper 1 of 2018: Language, India, States and Union Territories
  (Table C-16)},'' Government of India, New Delhi, 2018.

\bibitem{joshi2020state}
P.~Joshi, S.~Santy, A.~Budhiraja, K.~Bali, and M.~Choudhury, ``The state and
  fate of linguistic diversity and inclusion in the {NLP} world,'' in
  \emph{Proc. 58th Annual Meeting of the Association for Computational
  Linguistics (ACL)}, 2020, pp. 6282--6293.

\bibitem{bird2020decolonising}
S.~Bird, ``Decolonising speech and language technology,'' in \emph{Proc. 28th
  Int. Conf. Computational Linguistics (COLING)}, 2020, pp. 3504--3519.

\bibitem{du2020ppocr}
Y.~Du, C.~Li, R.~Guo, X.~Yin, W.~Liu, J.~Zhou, Y.~Bai, Z.~Yu, Y.~Yang, Q.~Dang,
  and H.~Wang, ``{PP-OCR}: A practical ultra lightweight {OCR} system,''
  arXiv:2009.09941, 2020.

\bibitem{jaidedai2020easyocr}
{JaidedAI}, ``{EasyOCR}: Ready-to-use {OCR} with 80+ supported languages,''
  \url{https://github.com/JaidedAI/EasyOCR}, 2020.

\bibitem{gcv2026languages}
{Google Cloud}, ``{Cloud Vision API}: {OCR} language support,''
  \url{https://cloud.google.com/vision/docs/languages}, 2026, accessed Oct.
  2026.

\bibitem{aws2026textract}
{Amazon Web Services}, ``Amazon {Textract} developer guide: Supported
  languages,'' \url{https://docs.aws.amazon.com/textract/}, 2026, accessed Oct.
  2026.

\bibitem{azure2026languages}
{Microsoft}, ``{Azure AI Vision}: Language support for {OCR},''
  \url{https://learn.microsoft.com/azure/ai-services/computer-vision/language-support},
  2026, accessed Oct. 2026.

\bibitem{fayaz2026kashmiri}
S.~A. Fayaz, M.~M. Khaja, A.~S. Bhat, D.~Mansoor, A.~Thapa, and M.~Zaman, ``A
  baseline optical character recognition framework for printed {Kashmiri}
  {Nastaliq} text using deep learning,'' \emph{Acadlore Trans. AI and Machine
  Learning}, vol.~5, no.~2, pp. 119--137, 2026.

\bibitem{iqbal2026koshurv3}
F.~Iqbal, ``{Koshur OCR v3 (Beta Baseline Model)},'' Hugging Face model card,
  \url{https://huggingface.co/Faizaniqbal/Koshur-OCR-v3-checkpoints}, 2026.

\bibitem{malik2026ksocr}
H.~N. Malik, ``{600K-KS-OCR}: A large-scale synthetic dataset for optical
  character recognition in {Kashmiri} script,'' arXiv:2601.01088, 2026.

\bibitem{jahan2025kashmiri}
Z.~Jahan and {Padmavathi S.}, ``Deciphering {Kashmiri} script: Challenges and
  advances in character recognition,'' in \emph{Proc. 12th Int. Conf. Computing
  for Sustainable Global Development (INDIACom)}, 2025, pp. 1--7.

\bibitem{zhou2017east}
X.~Zhou, C.~Yao, H.~Wen, Y.~Wang, S.~Zhou, W.~He, and J.~Liang, ``{EAST}: An
  efficient and accurate scene text detector,'' in \emph{Proc. IEEE Conf.
  Computer Vision and Pattern Recognition (CVPR)}, 2017, pp. 2642--2651.

\bibitem{baek2019craft}
Y.~Baek, B.~Lee, D.~Han, S.~Yun, and H.~Lee, ``Character region awareness for
  text detection,'' in \emph{Proc. IEEE/CVF Conf. Computer Vision and Pattern
  Recognition (CVPR)}, 2019, pp. 9357--9366.

\bibitem{redmon2016yolo}
J.~Redmon, S.~Divvala, R.~Girshick, and A.~Farhadi, ``You only look once:
  Unified, real-time object detection,'' in \emph{Proc. IEEE Conf. Computer
  Vision and Pattern Recognition (CVPR)}, 2016, pp. 779--788.

\bibitem{jocher2023yolov8}
G.~Jocher, A.~Chaurasia, and J.~Qiu, ``{Ultralytics YOLOv8},'' Version 8.0.0,
  \url{https://github.com/ultralytics/ultralytics}, 2023.

\bibitem{graves2006ctc}
A.~Graves, S.~Fern{\'a}ndez, F.~Gomez, and J.~Schmidhuber, ``Connectionist
  temporal classification: Labelling unsegmented sequence data with recurrent
  neural networks,'' in \emph{Proc. 23rd Int. Conf. Machine Learning (ICML)},
  2006, pp. 369--376.

\bibitem{shi2017crnn}
B.~Shi, X.~Bai, and C.~Yao, ``An end-to-end trainable neural network for
  image-based sequence recognition and its application to scene text
  recognition,'' \emph{IEEE Trans. Pattern Anal. Mach. Intell.}, vol.~39,
  no.~11, pp. 2298--2304, 2017.

\bibitem{shi2019aster}
B.~Shi, M.~Yang, X.~Wang, P.~Lyu, C.~Yao, and X.~Bai, ``{ASTER}: An attentional
  scene text recognizer with flexible rectification,'' \emph{IEEE Trans.
  Pattern Anal. Mach. Intell.}, vol.~41, no.~9, pp. 2035--2048, 2019.

\bibitem{li2023trocr}
M.~Li, T.~Lv, J.~Chen, L.~Cui, Y.~Lu, D.~Florencio, C.~Zhang, Z.~Li, and
  F.~Wei, ``{TrOCR}: Transformer-based optical character recognition with
  pre-trained models,'' in \emph{Proc. AAAI Conf. Artificial Intelligence},
  vol.~37, no.~11, 2023, pp. {13\,094}--{13\,102}.

\bibitem{bautista2022parseq}
D.~Bautista and R.~Atienza, ``Scene text recognition with permuted
  autoregressive sequence models,'' in \emph{Proc. European Conf. Computer
  Vision (ECCV)}, 2022, pp. 178--196.

\bibitem{blecher2024nougat}
L.~Blecher, G.~Cucurull, T.~Scialom, and R.~Stojnic, ``Nougat: Neural optical
  understanding for academic documents,'' in \emph{Proc. Int. Conf. Learning
  Representations (ICLR)}, 2024.

\bibitem{lorigo2006arabic}
L.~M. Lorigo and V.~Govindaraju, ``Offline {Arabic} handwriting recognition: A
  survey,'' \emph{IEEE Trans. Pattern Anal. Mach. Intell.}, vol.~28, no.~5, pp.
  712--724, 2006.

\bibitem{naz2014urdu}
S.~Naz, K.~Hayat, M.~I. Razzak, M.~W. Anwar, S.~A. Madani, and S.~U. Khan,
  ``The optical character recognition of {Urdu}-like cursive scripts,''
  \emph{Pattern Recognition}, vol.~47, no.~3, pp. 1229--1248, 2014.

\bibitem{sabbour2013upti}
N.~Sabbour and F.~Shafait, ``A segmentation-free approach to {Arabic} and
  {Urdu} {OCR},'' in \emph{Proc. SPIE Document Recognition and Retrieval XX},
  vol. 8658, 2013, p. 86580N.

\bibitem{ulhasan2013urdu}
A.~Ul-Hasan, S.~B. Ahmed, F.~Rashid, F.~Shafait, and T.~M. Breuel, ``Offline
  printed {Urdu} {Nastaleeq} script recognition with bidirectional {LSTM}
  networks,'' in \emph{Proc. 12th Int. Conf. Document Analysis and Recognition
  (ICDAR)}, 2013, pp. 1061--1065.

\bibitem{rahman2023utrnet}
A.~Rahman, A.~Ghosh, and C.~Arora, ``{UTRNet}: High-resolution {Urdu} text
  recognition in printed documents,'' in \emph{Proc. 17th Int. Conf. Document
  Analysis and Recognition (ICDAR)}, ser. Lecture Notes in Computer
  Science.\hskip 1em plus 0.5em minus 0.4em\relax Springer, 2023, pp. 305--324.

\bibitem{bashir2022kashmiri}
M.~Bashir, V.~Goyal, and K.~J. Giri, ``Challenges in recognition of {Kashmiri}
  script,'' in \emph{Recent Innovations in Computing}, ser. Lecture Notes in
  Electrical Engineering.\hskip 1em plus 0.5em minus 0.4em\relax Springer
  Singapore, 2022, pp. 33--43.

\bibitem{kiessling2019kraken}
B.~Kiessling, ``Kraken - an universal text recognizer for the humanities,'' in
  \emph{Proc. Digital Humanities (DH)}, Utrecht, The Netherlands, 2019.

\bibitem{jaderberg2014synth}
M.~Jaderberg, K.~Simonyan, A.~Vedaldi, and A.~Zisserman, ``Synthetic data and
  artificial neural networks for natural scene text recognition,'' NIPS
  Workshop on Deep Learning, arXiv:1406.2227, 2014.

\bibitem{gupta2016synthtext}
A.~Gupta, A.~Vedaldi, and A.~Zisserman, ``Synthetic data for text localisation
  in natural images,'' in \emph{Proc. IEEE Conf. Computer Vision and Pattern
  Recognition (CVPR)}, 2016, pp. 2315--2324.

\bibitem{zha2023datacentric}
D.~Zha, Z.~P. Bhat, K.-H. Lai, F.~Yang, and X.~Hu, ``Data-centric {AI}:
  Perspectives and challenges,'' in \emph{Proc. SIAM Int. Conf. Data Mining
  (SDM)}, 2023, pp. 945--948.

\bibitem{gadre2023datacomp}
S.~Y. Gadre \emph{et~al.}, ``{DataComp}: In search of the next generation of
  multimodal datasets,'' in \emph{Advances in Neural Information Processing
  Systems (NeurIPS), Datasets and Benchmarks Track}, 2023.

\bibitem{raqm}
{HOST-Oman}, ``Raqm: A library for complex text layout,''
  \url{https://github.com/HOST-Oman/libraqm}, 2024.

\bibitem{harfbuzz}
B.~Esfahbod \emph{et~al.}, ``{HarfBuzz}: Text shaping engine,''
  \url{https://github.com/harfbuzz/harfbuzz}, 2024.

\bibitem{iqbal2026synth}
F.~Iqbal, ``{Koshur-OCR-Synth},'' Hugging Face dataset,
  \url{https://huggingface.co/datasets/Faizaniqbal/Koshur-OCR-Synth}, 2026.

\bibitem{otsu1979threshold}
N.~Otsu, ``A threshold selection method from gray-level histograms,''
  \emph{IEEE Trans. Systems, Man, and Cybernetics}, vol.~9, no.~1, pp. 62--66,
  1979.

\end{thebibliography}

\appendices
\section{Recognizer Vocabulary}
\label{app:vocab}

The output layer of KashmiriCRNN has 101 classes. Index~0 is the CTC blank, and indices 1 to 100 correspond to the logical Unicode characters observed in the training corpus. The vocabulary covers the Kashmiri base letters, the Kashmiri vowel signs and diacritics (including \emph{zabar}, \emph{zer}, \emph{pesh}, the palatalization mark and \emph{nukta} combinations), Western and Perso-Arabic digits, and the punctuation retained by the corpus cleaner. The complete character list is distributed with the model weights.

\end{document}